\documentclass[runningheads]{llncs}
\usepackage[T1]{fontenc}
\usepackage{graphicx}
\usepackage{graphicx}
\usepackage{graphicx}
\usepackage{url}
\usepackage{cancel}
\usepackage{hyperref}
\usepackage{tabularx,booktabs}
\usepackage{multirow}
\usepackage{tcolorbox}
\newcolumntype{Y}{>{\centering\arraybackslash}X}
\usepackage{mathtools}
\usepackage{multirow, multicol}
\usepackage{xcolor}
\usepackage{amsmath}  % For general math features
\usepackage{amssymb}  % For \mathbb and \mathcal
\usepackage{arydshln}
\usepackage{makecell}

\usepackage{graphicx}  % For \scalebox
\begin{document}
\title{From Majority to Minority: A Diffusion-based Augmentation for Underrepresented Groups in Skin Lesion Analysis}
\titlerunning{From Majority to Minority: A Diffusion-based Augmentation}
\author{Janet Wang\and
Yunsung Chung\and
Zhengming Ding\and Jihun Hamm}
%index{Wang, Janet}
%index{Chung, Yunsung}
%index{Ding, Zhengming}
%index{Hamm, Jihun}
%
\authorrunning{Wang et al.}
\institute{Tulane University\\
\email{\{swang47, ychung3, zding1, jhamm3\}@tulane.edu}}

\maketitle              % typeset the header of the contribution
\begin{abstract}
AI-based diagnoses have demonstrated dermatologist-level performance in classifying skin cancer. However, such systems are prone to under-performing when tested on data from minority groups that lack sufficient representation in the training sets. Although data collection and annotation offer the best means for promoting minority groups, these processes are costly and time-consuming. Prior works have suggested that data from majority groups may serve as a valuable information source to supplement the training of diagnostic tools for minority groups. In this work, we propose an effective diffusion-based augmentation framework that maximizes the use of rich information from majority groups to benefit minority groups. Using groups with different skin types as a case study, our results show that the proposed framework can generate synthetic images that improve diagnostic results for the minority groups, even when there is little or no reference data from these target groups. The practical value of our work is evident in medical imaging analysis, where under-diagnosis persists as a problem for certain groups due to insufficient representation. Our implementation detail is available at \url{https://github.com/janet-sw/skin-diff}.

\keywords{Skin Lesion Analysis  \and Diffusion Models \and Data Augmentation.}
\end{abstract}
%
%
% ********************************************************************
\section{Introduction}
AI-assisted diagnostic systems demonstrate expert-level capability in classifying skin cancers, often identified visually \cite{Esteva2017DermatologistlevelCO,Liu2020-vx,Brinkerarticle}. These systems can potentially contribute to teledermatology as diagnostic and decision-support tools, enhancing diagnostic accessibility in rural areas \cite{Coustasse2019UseOT}. However, despite such success, recent studies have highlighted their susceptibility to under-diagnosing minority groups, such as those with underrepresented skin types, hindering their ability to generalize across different demographic groups \cite{daneshjou2022disparities,groh2021evaluating}. Although the majority group contains rich lesion information, directly training models for cross-color classification using this data is challenging due to the domain gap caused by varying skin types \cite{wang2024achieving}. Prior research has suggested using synthetic images generated from majority groups to supplement the training of AI for minority groups \cite{rezk2022improving}.

Augmenting skin condition data with synthetic images has been explored, owing to its potential to address common challenges for skin lesion analysis, such as data privacy, imbalance, and scarcity. Notably, Generative Adversarial Networks (GANs) \cite{goodfellow2020generative} and Diffusion Models (DMs) \cite{dhariwal2021diffusion} have emerged as leading techniques for generating high-quality skin lesion images. While GANs have successfully produced photorealistic synthetic images, their generation is uncontrollable \cite{pmlr-v116-ghorbani20a,qin2020gan}. On the other hand, DMs pre-trained on extensive web data have enabled higher controllability over image generation through the guidance of textual prompts, allowing for the creation of diverse and high-fidelity images of target skin conditions and skin types.

Existing studies have tried diffusion models to augment minority skin types using two public datasets: Diverse Dermatology Images (DDI) \cite{daneshjou2022disparities} and Fitzpatrick17k \cite{groh2021evaluating}. Each image in these datasets is annotated with skin type labels based on the Fitzpatrick scoring system \cite{fitzpatrick1988validity}. In their work, \cite{sagers2023augmenting} generated multiple synthetic images for each real image using Stable Diffusion \cite{Rombach_2022_CVPR} and then trained the classifier on a dataset including real and synthetic data. They found that diffusion models can enhance diagnosis accuracy across skin types in binary malignancy classification on the DDI dataset, though the number of real images is the key driver in performance. Additionally, \cite{sagers2022improving} sampled a small number of seed images with skin types at the ends of the Fitzpatrick spectrum (FST I-II and FST V-VI) and carefully cropped the disease pathology in them, before generating synthetic data from the seeds using OpenAI DALL·E 2's inpainting feature. They conducted class-wise data augmentation by incorporating synthetic images of the target condition and minority skin type into the real training set. Other related studies have focused on internal datasets \cite{akrout2023diffusionbased,ktena2023generative}. 

\begin{figure}[!tbp]
\centering
\includegraphics[width=1.0 \textwidth]{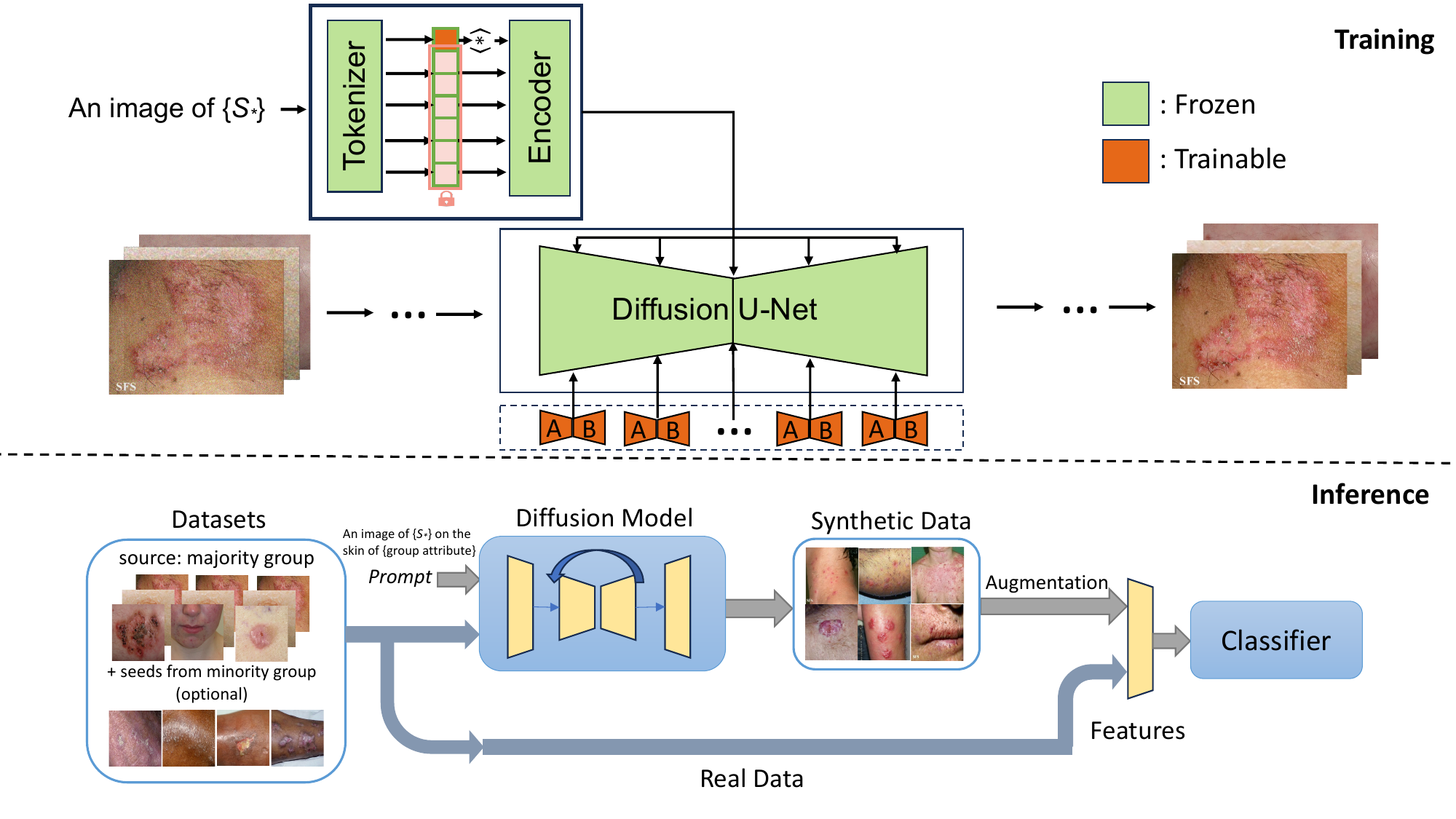}
\caption{Overview of the proposed augmentation framework. The framework pairs each training image with a textual prompt describing the condition as an input to train a latent diffusion model. Embeddings associated with new lesion concepts are found through Textual Inversion. Compact matrices $A$ and $B$ are optimized via LoRA to facilitate training with the new embeddings. During inference, the trained model produces synthetic images from the training set that mainly features the majority groups via image-to-image generation, thus conditioned on visual cues of lesions from images and textual prompts describing the target condition and group attributes.}
\label{fig:illustration}
\end{figure}

Despite these advancements, the potential to leverage diffusion models' knowledge about skin variation and the rich lesion information from majority groups to benefit minority groups remains underexplored. In this work, we propose a novel diffusion-based augmentation framework capable of learning skin lesion concepts from majority groups and generating images to improve classification performance for minority groups. Unlike current works that assume the existence of data from minority groups, we hypothesize that the information gained from majority groups and the diffusion model's pre-trained knowledge is sufficient to generate useful synthetic data. We test our hypothesis in a challenging multi-condition classification task. The framework is illustrated in Fig. \ref{fig:illustration}. We conduct our experiments on the Fitzpatrick17k dataset, which includes lesions that are less familiar to diffusion models than common skin cancer. This dataset has a skewed skin type distribution, with light skin types (FST I-II) being significantly more than dark skin types (FST V-VI), thus forming majority and minority groups. Our investigation focuses on images from both groups and is structured around three scenarios with increasing difficulty: \textbf{(i)} the training source includes some data from both groups; \textbf{(ii)} there is limited data from the minority group in the training source; and \textbf{(iii)} the training source lacks data from the minority group. Through extensive experiments and analysis, we found that:

\begin{itemize}
    \item Our proposed method effectively leverages lesion information from the majority group to generate synthetic images that can improve classification for the minority group across all settings, even without reference data from the minority group.
    \item Using synthetic images generated by our method to train classifiers consistently outperforms training with real images across various architectures. Further improvement is observed when combining real and synthetic data.
    \item Our method is sensitive to information from the minority group. A notable improvement can be observed when even a few examples from the minority group are added to the training set.
\end{itemize}
%
%
% ********************************************************************
\section{Methods}
\label{sec:methods}
In this section, we will introduce key techniques that have been adapted for skin disease datasets in our proposed augmentation framework. 

\begin{table*}[ht]
\centering
\resizebox{1.0 \textwidth}{!}{%
\begin{tabular}{@{}ccccccccc@{}}
\toprule
\textit{\textbf{FST}} & \textbf{\begin{tabular}[c]{@{}c@{}}Basal Cell \\ Carcinoma\end{tabular}} & \textbf{Folliculitis} & \textbf{\begin{tabular}[c]{@{}c@{}}Nematode \\ Infection\end{tabular}} & \textbf{\begin{tabular}[c]{@{}c@{}}Neutrophilic \\ Dermatoses\end{tabular}} & \textbf{\begin{tabular}[c]{@{}c@{}}Prurigo \\ Nodularis\end{tabular}} & \textbf{Psoriasis} & \textbf{\begin{tabular}[c]{@{}c@{}}Squamous Cell \\ Carcinoma\end{tabular}} & \textit{\textbf{Total}} \\ \midrule

\textit{I} & 85 & 30 & 15 & 70 & 7 & 113 & 100 & 420 \\

\textit{II} & 156 & 97 & 56 & 115 & 28 & 232 & 180 & 864 \\

\textit{V} & 24 & 31 & 32 & 31 & 29 & 64 & 40 & 251 \\

\textit{VI} & 7 & 9 & 12 & 15 & 9 & 21 & 23 & 96 \\

Total & 272       & 167          & 115       & 231          & 73       & 430       & 343           & \textbf{1631} \\ \bottomrule
\end{tabular}%
}
\caption{Sample distribution across skin conditions by Fitzpatrick Skin Type.}
\label{tab:dataset_table}
\end{table*}
\subsubsection{Latent Diffusion Models} We implement our method using Latent Diffusion Models (LDMs) \cite{Rombach_2022_CVPR}, a class of Denoising Diffusion Probabilistic Models (DDPMs) \cite{ho2020denoising} that operate in the latent space of an autoencoder, to enable DDPM training with limited computational resources. LDMs include two core components: a pre-trained autoencoder and a diffusion model. In our study, the encoder of the autoencoder $\mathcal{E}$ encodes skin lesion images $x\in \mathcal{D}_x$ into a latent representation $z = \mathcal{E}(x)$, while the decoder $D$ maps the latent representations back to images, such that $D\left(\mathcal{E}(x)\right)\approx x$. The diffusion model is trained to generate representations conditioned on prompts describing skin disease and skin type, within the learned latent space. Let $c_\theta(y)$ be a model that maps a conditioning input $y$ into a vector. We then learn the conditional LDM via
\begin{equation}
\scalebox{1.0}{$
    L_{LDM} \coloneqq \mathbb{E}_{z\in\mathcal{E}(x), ~y,~ \epsilon \in \mathcal{N}(0, 1),~ t}
    \Big[ \Vert \epsilon - \epsilon_\theta(z_{t},t, c_\theta(y)) \Vert_{2}^{2}\Big] \, ,
$}
\label{eq:LDM_loss}
\end{equation}
where $t$ is the time step, $z_t$ is the latent noise at time $t$, $\epsilon$ is the unscaled noise sample, and $\epsilon_\theta$ is the denoising network.
\subsubsection{Concept Discovery via Textual Inversion} Our proposed framework leverages Textual Inversion \cite{gal2023an} to capture a unique embedding that accurately represents the targeted skin lesion concept from training data. Skin lesion images paired with a string containing a placeholder word (e.g., \lq{An image of $\{S_{*}\}$}\rq) are used to guide the learning of a new lesion embedding for the generative model. In particular, the optimal embedding $v_*$ that encapsulates the lesion concept $S_{*}$ is derived by minimizing the reconstruction loss, 
\begin{equation}
\scalebox{1.0}{$
    v_* = \mathrm{argmin}_v 
    \mathbb{E}_{z\sim\mathcal{E}(x), y, \epsilon \sim \mathcal{N}(0, 1), t }\Big[ \Vert \epsilon - \epsilon_\theta(z_{t},t, c_\theta(y, S_*)) \Vert_{2}^{2}\Big] \, ,
$}
\label{eq:v_opt}
\end{equation}
where the same training scheme as the original LDM model is used, with $c_\theta$ and $\epsilon_\theta$ fixed. 
\subsubsection{Fine-grained Detail Enhancement with LoRA} To enhance efficiency in fine-tuning LDM, we employ Low-Rank Adaptation (LoRA) \cite{hu2022lora} in our framework, with the discovered tokens after textual inversion. This fine-tuning strategy freezes the pre-trained model weights and introduces two compact matrices $A$ and $B$, where $A  \in \mathbb{R}^{n\times r}, B \in \mathbb{R}^{r\times n}$. The adaptation matrices $AB$ are integrated into the attention layers to capture fine visual details of the skin lesion that were not initially present in the pre-trained model, with target embedding $v_*$. The optimization is formulated as
\begin{equation}
\scalebox{1.0}{$
    L := \mathbb{E}_{z\sim\mathcal{E}(x), y, \epsilon \sim \mathcal{N}(0, 1), t }\Big[ \Vert \epsilon - \epsilon_{\theta_{AB}}(z_{t},t, c_{\theta_{AB}}(y, v_*)) \Vert_{2}^{2}\Big] \, .
    % \label{eq:FINAL_loss}
$}
\end{equation}
\begin{table*}[!tbp]
\centering
% \resizebox{\columnwidth}{\textwidth}{!}{%
\resizebox{1.0 \textwidth}{!}{%
\begin{tabular}{cllrrrr}
\Xhline{1pt}
\multicolumn{1}{l}{Architecture} &
  Train Type&
  Train Size&
  \multicolumn{1}{l}{Accuracy} &
  \multicolumn{1}{l}{Precision} &
  \multicolumn{1}{l}{Recall} &
  \multicolumn{1}{l}{F1 Score} \\ \hline
\multirow{3}{*}{VGG-16}    & real        & 1519 & 70.24 $\pm$ 0.12 & 72.49 $\pm$ 0.37 & 69.58 $\pm$ 0.30 & 70.48 $\pm$ 0.13 \\ 
                           & syn         & 1519 * 5 & 75.00 $\pm$ 0.64 & 75.77 $\pm$ 0.39 & 73.17 $\pm$ 0.29 & 72.42 $\pm$ 0.61 \\  
                           & real + syn  & 1519 * 6 & 77.98 $\pm$ 0.40 & 81.51 $\pm$ 0.25 & 78.87 $\pm$ 0.23 & 77.45 $\pm$ 0.29 \\ \hline
\multirow{3}{*}{ResNet-18} & real        & 1519 & 68.45 $\pm$ 0,42 & 69.42 $\pm$ 0.57 & 69.05 $\pm$ 0.42 & 67.94 $\pm$ 0.35 \\ 
                           & syn         & 1519 * 5 & 69.05 $\pm$ 0.36 & 69.57 $\pm$ 0.82 & 69.05 $\pm$ 0.42 & 68.02 $\pm$ 0.33 \\
                           & real + syn  & 1519 * 6 & 71.36 $\pm$ 0.69 & 69.02 $\pm$ 0.48 & 68.45 $\pm$ 0.40 & 67.56 $\pm$ 0.59 \\ \hline
\multirow{3}{*}{ViT-B-16}  & real        & 1519 & 70.38 $\pm$ 0.42 & 73.72 $\pm$ 0.69 & 70.82 $\pm$ 0.39 & 70.61 $\pm$ 0.59 \\ 
                           & syn         & 1519 * 5 & 74.19 $\pm$ 0.37 & 77.89 $\pm$ 1.01 & 74.04 $\pm$ 0.85 & 73.58 $\pm$ 0.53 \\
                           & real + syn  & 1519 * 6 & 78.65 $\pm$ 0.53 & 81.57 $\pm$ 0.47 & 79.17 $\pm$ 0.84 & 78.24 $\pm$ 0.64 \\ \Xhline{1pt}
\end{tabular}%
}
\caption{This table presents the results when the training set includes non-flexible images from the minority group (291 of FST V-VI) and the majority group (1228 FST I-II). The test set is a flexible subset of the minority group (56 of FST V-VI), uniformly distributed across the 7 conditions. Here, ``real" indicates that the classifier is trained solely on real images, while ``syn" means that it is trained exclusively on synthetic images generated by our framework. Accordingly, ``real+syn" means the subsequent classifier is trained on a combination of both.
}
\label{table_1}
\end{table*}
\section{Experiments}
We conduct our experiments using the Fitzpatrick17k dataset, where each image is annotated with a condition and a Fitzpatrick Skin Type (FST) label. In line with \cite{sagers2022improving}, our analysis narrows down to a subset of the Fitzpatrick17k dataset, encompassing 7 conditions (Table \ref{tab:dataset_table}). These conditions were selected because they represent the largest sample sizes at the ends of the Fitzpatrick Skin Type (FST I-II or V-VI) spectrum. Unlike \cite{sagers2022improving}, our study excludes intermediate skin types (FST III-IV), to explore the efficacy of our diffusion-based augmentation in a more challenging and explainable way. We randomly sample 8 images for each condition from the lightest (FST I-II) and darkest (FST V-VI) skin type groups, resulting in a \textbf{flexible subset} of 56 images for each group.

We examine three scenarios: 
\textbf{(i)} the training set includes images of dark and light skin types, and the test set features the uniformly distributed flexible subset across the 7 conditions; \textbf{(ii)} the training set predominantly includes light-skinned images and a few dark-skinned images, while the test set consists of dark-skin data; \textbf{(iii)} the training set lacks dark-skinned images entirely, while the test set comprises dark-skinned images. In all scenarios, we generate 5 synthetic images for each real one in the training set during inference, using the fine-tuned model, as illustrated in Fig. \ref{fig:illustration}. In scenario \textbf{(i)}, to ensure a sufficient number of examples for both majority and minority groups in the training set, we designate the flexible subset of dark skin as the test set and use remaining non-flexible images for generator and classifier training. This setting also serves as the basis for hyperparameter tuning of the diffusion model, with the selected hyperparameters being fixed for subsequent experiments.

\textbf{Implementation Details} In each setting, we randomly sampled 5 flexible subsets and repeated the experiment 5 times. We used the Stable Diffusion 2.1 base \cite{Rombach_2022_CVPR} and the Diffusers library \cite{von-platen-etal-2022-diffusers} for fine-tuning the diffusion model and generating synthetic images. For classifier backbones, we utilized pre-trained VGG-16 \cite{simonyan2014very}, ResNet-18 \cite{he2016deep}, and ViT-B-16 \cite{wu2020visual} and trained each classifier using the Adam optimizer with an initial learning rate of 1e-3 and transformations as in \cite{groh2021evaluating}. A weight-based sampler and StepLR scheduler were applied. All experiments were conducted on two NVIDIA GeForce RTX 3090s.
%
%
% ********************************************************************
\begin{table*}[!tbp]
\centering
% \resizebox{\columnwidth}{!}{%
\resizebox{1.0 \textwidth}{!}{%
\begin{tabular}{cllrrrr}
\Xhline{1pt}
\multicolumn{1}{l}{Architecture} &
  Train Type&
  Train Size&
  \multicolumn{1}{l}{Accuracy} &
  \multicolumn{1}{l}{Precision} &
  \multicolumn{1}{l}{Recall} &
  \multicolumn{1}{l}{F1 Score} \\ \hline
\multirow{3}{*}{VGG-16}    & real        & 1284 & 58.79 $\pm$ 0.10 & 58.90 $\pm$ 0.03 & 58.26 $\pm$ 0.05 & 56.98 $\pm$ 0.04 \\ 
                           & syn         & 1284 * 5 & 62.86 $\pm$ 0.15 & 61.49 $\pm$ 0.15 & 63.32 $\pm$ 0.13 & 61.57 $\pm$ 0.24 \\  
                           & real + syn  & 1284 * 6 & 63.66 $\pm$ 0.11 & 62.80 $\pm$ 0.12 & 64.08 $\pm$ 0.08 & 62.72 $\pm$ 0.18 \\ \hline
\multirow{3}{*}{ResNet-18} & real        & 1284 & 50.31 $\pm$ 0.30 & 50.45 $\pm$ 0.44 & 51.27 $\pm$ 0.30 & 48.23 $\pm$ 0.32 \\ 
                           & syn         & 1284 * 5 & 56.36 $\pm$ 0.16 & 56.58 $\pm$ 0.13 & 59.78 $\pm$ 0.08 & 55.58 $\pm$ 0.10 \\ 
                           & real + syn  & 1284 * 6 & 61.33 $\pm$ 0.17 & 59.75 $\pm$ 0,12 & 62.52 $\pm$ 0.17 & 59.94 $\pm$ 0.17 \\ \hline
\multirow{3}{*}{ViT-B-16}  & real        & 1284 & 62.03 $\pm$ 0.19 & 63.09 $\pm$ 0.13 & 62.02 $\pm$ 0.04 & 61.05 $\pm$ 0.04 \\ 
                           & syn         & 1284 * 5 & 68.83 $\pm$ 0.19 & 70.07 $\pm$ 0.03 & 68.22 $\pm$ 0.25 & 68.34 $\pm$ 0.19 \\ 
                           & real + syn  & 1284 * 6 & 71.20 $\pm$ 0.07 & 71.61 $\pm$ 0.19 & 71.53 $\pm$ 0.01 & 71.17 $\pm$ 0.13 \\ \Xhline{1pt}
\end{tabular}%
}
\caption{Classification results when the training set contains a few reference images from the dark-skinned flexible subset (56 of FST V-VI) and non-flexible light-skinned images (1228 of FST I-II). The test set includes all other dark-skinned images (291 of FST V-VI) outside the flexible subset.
}
 \label{table_2}
\end{table*}
%
%
%
% ********************************************************************
\section{Results}
To assess the efficacy of our augmentation framework across the three settings, we train the classifier on data that includes real images only, synthetic images only, or a combination of both, respectively. Our evaluation is based on four metrics: accuracy, precision, recall, and F1. First, in the setting with some images from both majority and minority groups in the training set, we observe that synthetic data enhances performance across all architectures (Table \ref{table_1}). Specifically, classifiers trained on synthetic images consistently outperform those trained solely on real ones, and the combination of both types of data for training yields further improvements. 

\begin{figure*}[t]
  % \raggedleft % Aligns the image to the left
   \centering
   \includegraphics[width=1.0 \linewidth]{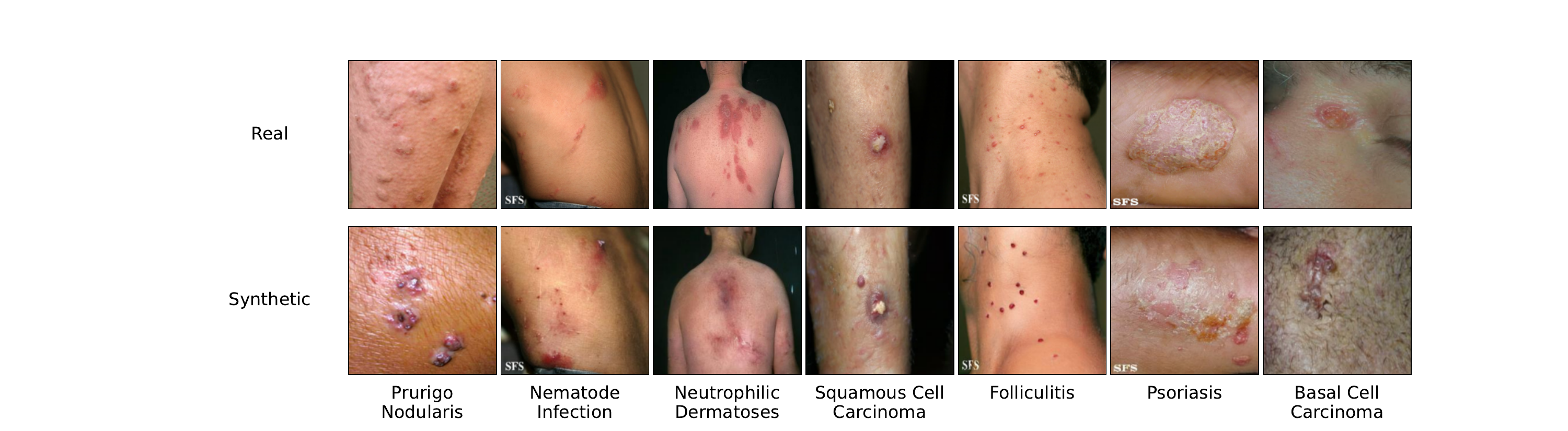}
   \caption{Examples of synthetic images generated by a model trained exclusively on light-skinned images, using prompts describing dark skin types.}
   \label{fig:demo_img}
\end{figure*}

This trend of consistent improvement is also evident in the more challenging scenarios where there are little or no reference images from the minority groups (Tables \ref{table_2} and \ref{table_3}) in the training set. Notably, significant improvement is observed when just a few reference images from the minority group are available in the training set for image generation and classification. The transformer-based classifier demonstrates a larger improvement gap over the real image baseline than the CNN-based models. In the most challenging setting, with no reference images from the minority group, the improvement margin narrowed, suggesting that our pipeline effectively maximizes the use of limited information from the flexible subset of the minority group during training. Despite these challenges, the sustained improvements in the third setting validate our framework's effectiveness in transferring information across groups. Examples of real and synthetic image pairs for each condition are presented in Fig. \ref{fig:demo_img}. Qualitatively, the synthetic images generated by our augmentation framework introduce more diversity to the training sets, including variations in skin color and lesion patterns.
% ********************************************************************

\begin{table*}[!tbp]
\centering
% \resizebox{\columnwidth}{!}{%
\resizebox{1.0 \textwidth}{!}{%
\begin{tabular}{cllrrrr}
\Xhline{1pt}
\multicolumn{1}{l}{Architecture} &
  Train Type&
  Train Size&
  \multicolumn{1}{l}{Accuracy} &
  \multicolumn{1}{l}{Precision} &
  \multicolumn{1}{l}{Recall} &
  \multicolumn{1}{l}{F1 Score} \\ \hline
\multirow{3}{*}{VGG-16}    & real         & 1228 & 55.58 $\pm$ 0.10 & 54.60 $\pm$ 0.11 & 51.84 $\pm$ 0.29 & 51.97 $\pm$ 0.45 \\ 
                           & syn          & 1228 * 5 & 57.62 $\pm$ 0.09 & 56.62 $\pm$ 0.15 & 55.42 $\pm$ 0.20 & 55.36 $\pm$ 0.22 \\  
                           & real + syn   & 1228 * 6 & 58.08 $\pm$ 0.08 & 57.36 $\pm$ 0.13 & 55.78 $\pm$ 0.17 & 55.85 $\pm$ 0.23 \\ \hline
\multirow{3}{*}{ResNet-18} & real         & 1228 & 49.42 $\pm$ 0.36 & 49.79 $\pm$ 0.44 & 48.30 $\pm$ 0.23 & 47.42 $\pm$ 0.30 \\ 
                           & syn          & 1228 * 5 & 53.47 $\pm$ 0.32 & 52.39 $\pm$ 0.30 & 53.79 $\pm$ 0.30 & 51.97 $\pm$ 0.29 \\ 
                           & real + syn   & 1228 * 6 & 56.50 $\pm$ 0.15 & 55.72 $\pm$ 0.15 & 55.10 $\pm$ 0.16 & 54.76 $\pm$ 0.18 \\ \hline
\multirow{3}{*}{ViT-B-16}  & real         & 1228 & 57.66 $\pm$ 0.35 & 61.89 $\pm$ 0.68 & 55.45 $\pm$ 0.36 & 55.96 $\pm$ 0.51 \\ 
                           & syn          & 1228 * 5 & 58.94 $\pm$ 0.23 & 62.79 $\pm$ 0.20 & 55.62 $\pm$ 0.22 & 56.53 $\pm$ 0.18 \\ 
                           & real + syn   & 1228 * 6 & 60.96 $\pm$ 0.31 & 64.57 $\pm$ 0.01 & 57.51 $\pm$ 0.29 & 58.92 $\pm$ 0.31 \\ \Xhline{1pt}
\end{tabular}%
}
\caption{Classification results when no image from the minority group is in the training set, which only has non-flexible images of the majority group (1228 of FST I-II). The test set has non-flexible images of the minority group (291 of FST V-VI).
}
\label{table_3}
\end{table*}

To further investigate our framework's generation capabilities, we conduct an ablation study comparing our framework with various generation strategies. This study focuses on the first setting, where the test set consists of light- or dark-skinned flexible images (56 for each type). We first examine the Stable Diffusion's vanilla text-to-image and image-to-image pipelines to generate synthetic images. Next, we leverage Textual Inversion to learn the lesion embeddings and then generate synthetic images from these embeddings, with text-to-image and image-to-image pipelines. Since image-to-image outperforms text-to-image in both vanilla SD and Textual Inversion generation, we focus on image-to-image generation after fine-tuning the diffusion model using LoRA, to investigate if optimizing the diffusion model can benefit the generation even more. We train a VGG-16 using only the synthetic images and then compare these generation strategies with ours, as shown in Table \ref{tab:ablation}. 

Overall, training classifiers on synthetic images generated by text-to-image models proves less effective than employing image-to-image techniques, underscoring the importance of visual cues in augmenting skin lesion classification. Additionally, using off-the-shelf models for image generation yields less improvement than training strategies such as Textual Inversion and LoRA, regardless of whether the target is a minority or majority group. Finally, the combination of Textual Inversion and LoRA results in the highest accuracy, thereby validating the practicality of our design which integrates these two strategies. This improvement can be explained by the model's enhanced ability to associate the fine visual cues of the lesion with the learned textual tokens. 

Since a direct comparison with existing related works is challenging due to the uncertain use of data, this ablation study can serve as an indirect comparison. As introduced previously, related works leveraged off-the-shelf diffusion models such as DALL·E or fine-tuned a Stable Diffusion model for text-to-image generation. Our results demonstrate that utilizing the dual guidance of visual cues and text prompts via fine-tuning diffusion models can maximize the potential of diffusion-based augmentation and enhance the diagnosis for minority groups.

\begin{table}[t]
\centering
\resizebox{0.75 \columnwidth}{!}{%
\begin{tabular}{ccc|cccc}
\Xhline{1pt}
           & \multicolumn{2}{c|}{txt2img} & \multicolumn{4}{c}{img2img}       \\ \cdashline{2-7}[1pt/1pt]
test     & vanilla       & ti          & vanilla & ti    & lora  & ti+lora \\ \hline
light (56) & 18.80         & 35.36       & 48.21   & 46.43 & 52.00 & 53.57   \\
dark (56)  & 21.22         & 44.64       & 69.64   & 71.43 & 73.21 & 79.57   \\ 
\Xhline{1pt}
\end{tabular}%
}
\caption{Classification accuracy with various generation strategies for the first setting. Here, ``vanilla" stands for Stable Diffusion's original text-to-image and image-to-image pipelines, ``ti" for Textual Inversion, and ``lora" for LoRA.}
\label{tab:ablation}
\end{table}

\section{Conclusion}
In this work, we present an effective diffusion-based augmentation framework that consistently improves classification results for the minority group, even when training the classifier exclusively with synthetic images. This improvement is observed regardless of the availability of reference data from the minority group in the training set. The ablation study also validates that our framework's dual-guidance generation approach successfully learns novel lesion concepts previously unfamiliar to the diffusion models. A practical takeaway from this study is that, even in cases of data scarcity, existing data and diffusion models can still provide valuable insights, maximizing information usage and achieving better performance. In the future, we plan to apply this technique to other medical datasets characterized by significant differences in group sizes. Additionally, as we used all synthetic images generated for each setting without any filtering mechanism, we also aim to investigate which types of synthetic data are useful for lesion diagnosis and how to generate them.

\begin{credits}
\subsubsection{\ackname} This work was partly supported by the NSF EPSCoR-Louisiana Materials Design Alliance (LAMDA) program \#OIA-1946231 and partly by the Harold L. and Heather E. Jurist Center of Excellence for Artificial Intelligence at Tulane University.

\end{credits}
\bibliographystyle{splncs04}
\bibliography{Paper-0010}

% \end{thebibliography}
\end{document}